\documentclass[letterpaper]{article}
\usepackage{proceed2e}
\usepackage[margin=1in]{geometry}

\usepackage{graphicx,xcolor}
\usepackage{colortbl}
\usepackage{booktabs}
\usepackage{algorithm}
\usepackage{algorithmicx}
\usepackage[noend]{algpseudocode}
\usepackage{amsmath,amsfonts,amssymb}
\usepackage{cancel}
\usepackage{microtype}
\usepackage{bbm}

\usepackage[]{natbib}
    
\def\ci{\perp\!\!\!\!\perp}

\title{Conditional independence testing based on a nearest-neighbor estimator of conditional mutual information}

\author{ {\bf Jakob Runge\thanks{Also at: German Aerospace Agency, Institute of Data Science, Jena, Germany.}} \\
Grantham Institute, Imperial College London,\\
SW72AZ London, United Kingdom
}

\begin{document}

\maketitle

\begin{abstract}
Conditional independence testing is a fundamental problem underlying causal discovery and a particularly challenging task in the presence of nonlinear and high-dimensional dependencies.
Here a fully non-parametric test for continuous data based on conditional mutual information combined with a local permutation scheme is presented. Through a nearest neighbor approach, the test efficiently adapts also to non-smooth distributions due to strongly nonlinear dependencies. Numerical experiments demonstrate that the test reliably simulates the null distribution even for small sample sizes and with high-dimensional conditioning sets.
The test is better calibrated than kernel-based tests utilizing an analytical approximation of the null distribution, especially for non-smooth densities, and reaches the same or higher power levels. Combining the local permutation scheme with the kernel tests leads to better calibration, but suffers in power.
For smaller sample sizes and lower dimensions, the test is faster than random fourier feature-based kernel tests if the permutation scheme is (embarrassingly) parallelized, but the runtime increases more sharply with sample size and dimensionality. Thus, more theoretical research to analytically approximate the null distribution and speed up the estimation for larger sample sizes is desirable. 
\end{abstract}

\section{Introduction}
Conditional independence testing lies at the heart of causal discovery \citep{Spirtes2000} and at the same time is one of its most challenging tasks. For observed random variables $X, Y, Z$, measuring that $X$ and $Y$ are independent given $Z$, denoted as $X \ci Y | Z$, implies that no causal link can exist between $X$ and $Y$ under the relatively weak assumption of \emph{faithfulness} \citep{Spirtes2000}. A finding of conditional independence is then more pertinent to causal discovery than a finding of (conditional) \emph{dependence} from which a causal link only follows under stronger assumptions \citep{Spirtes2000}.

Here we focus on the difficult case of continuous variables \citep{Bergsma2004}. While various conditional independence (CI) tests exist if assumptions such as linearity or additivity \citep{Daudin1980a,Peters2014} are justified (for a numerical comparison see \citet{Ramsey2014}), here we focus on the general definition of CI implying that the conditional joint density factorizes: $p(X,Y|Z)=p(X|Z) p(Y|Z)$. Note that wrong assumptions can lead to incorrectly detecting CI (type II error, false negative), but also to wrongly concluding on conditional dependence (type I error, false positive). 

Recent research has focused on the general case without assuming a functional form of the dependencies as well as the data distributions. One approach is to discretize the variable $Z$ and make use of easier unconditional independence tests $X\ci Y | Z=z$ \citep{Margaritis2005,Huang2010}. 
However, this method suffers from the curse of dimensionality for high-dimensional conditioning sets $Z$. 

On the other hand, kernel-based methods are known for their capability to deal with nonlinearity and high dimensions \citep{Fukumizu2008}. A popular test is the Kernel Conditional Independence Test (KCIT) \citep{Zhang2012} which essentially tests for zero Hilbert-Schmidt norm of the partial cross-covariance operator, or the Permutation CI test \citep{Doran2014} which solves an optimization problem to generate a permutation surrogate on which kernel-two sample testing can be applied. 
Kernel methods suffer from high computational complexity since large kernel matrices have to be computed.
\citet{Strobl2017} present two types of orders of magnitude faster CI tests based on approximating kernel methods using \emph{random Fourier features}, called Randomized Conditional Correlation Test (RCoT) and Randomized Conditional Independence Test (RCIT). RCoT can be related to kernelized two-step conditional independence testing \citep{Zhang2017}. Last, \citet{Wang2015} proposed a \emph{conditional distance correlation} (CDC) test based on the correlation of distance matrices between $X,Y,Z$ which have been linked to kernel-based approaches \citep{Sejdinovic2013}.

Kernel methods in general require carefully adjusted bandwidth parameters that characterize the length scales between samples in the different subspaces of $X,Y,Z$. These bandwidths are \emph{global} in each subspace in the sense that they are applied on the whole range of values for $X,Y,Z$, respectively. 

Testing for independence requires access to the null distribution under CI. \citet{Strobl2017} and \citet{Wang2015} derived asymptotic approximations of the theoretical null distributions, but such approximations only hold for larger sample sizes. 
The alternative are permutation-based approaches, where the null-distribution is generated by computing the test statistic from a permuted sample.

Our approach to testing CI is founded in an information-theoretic framework. The conditional mutual information (CMI) is zero if and only if $X \ci Y | Z$. While some kernel-based measures can also be related to information-theoretic quantities (see, e.g., \citet{Fukumizu2008}), our approach is to \emph{directly} estimate CMI by combining the well-established Kozachenko-Leonenko $k$-nearest neighbor estimator \citep{kozachenko1987sample,Kraskov2004a,FrenzelPompe2007,Vejmelka2008,Poczos2012b,Gao2017} with a nearest-neighbor local permutation scheme. Their main advantage is that nearest-neighbor statistics are \emph{locally adaptive}: The hypercubes around each sample point are smaller where more samples are available. Unfortunately, few theoretical results are available for the complex mutual information estimator. While the Kozachenko-Leonenko estimator is asymptotically unbiased and consistent \citep{kozachenko1987sample,Leonenko2008a}, the variance and finite sample convergence rates are unknown. Hence, our approach relies on a local permutation test that is also based on nearest neighbors and, hence, data-adaptive.

Our numerical experiments comparing the CMI test with KCIT, RCIT, RCoT, and CDC show that the test reliably simulates the null distribution even for small sample sizes and with high dimensional conditioning sets. 
The local permutation scheme yields better calibrated tests for sample sizes below 1000 than kernel-based tests relying on asymptotic approximations such as KCIT, RCIT or RCoT. We also tested RCIT and RCoT combined with our local permutation scheme which yields better calibration for smaller sample sizes. 
The CMI test reaches the same or higher power levels than the other compared approaches, especially for highly nonlinear dependencies.
The computational time of both the CMI test and the kernel tests strongly depends on hyperparameters. We found that for smaller samples sizes CMI is faster than RCIT or RCoT by making use of KD-tree neighbor search methods \citep{Maneewongvatana1999}, but a major drawback is its computationally expensive permutation scheme making more theoretical research to analytically approximate the null distribution for larger sample sizes desirable. Code for the CMI test is freely available at \texttt{https://github.com/jakobrunge/tigramite}.

\section{Conditional independence test}

\subsection{Conditional mutual information}

CMI for continuous and possibly multivariate random variables $X,Y,Z$ is defined as 
\begin{align}
& I_{X;Y  | Z } \nonumber\\ 
&=   \iiint dx dy dz~  p(x,y,z) \log \frac{ p(x,y |z)}{p(x |z)\cdot p(y |z)}\\
& = H_{XZ} + H_{YZ} - H_{Z} - H_{XYZ} \label{eq:def_cmi} \,,
\end{align}
where $H$ denotes the Shannon entropy and where we have to assume that the densities $p(\cdot)$ exist. We wish to test the conditional independence hypothesis
\begin{align}
H_0 : ~~ X ~&\ci ~Y~ |~ Z 
\end{align}
versus the general alternative.
From the definition of CMI it is immediately clear that $I_{X;Y| Z}=0$ if and only if $X \ci Y | Z$, provided that the densities are well-defined. 
Shannon-type conditional mutual information is theoretically well-founded and its value is well interpretable as the shared information between $X$ and $Y$ not contained in $Z$. While this does not immediately matter for a conditional independence test's $p$-value, causal discovery algorithms often make use of the test statistic's value, for example to sort the order in which conditions are tested. CMI here readily allows for an interpretation in terms of the relative importance of one condition over another. Note that the test statistic values of kernel-based tests typically depend on the chosen kernel.

\subsection{Nearest-neighbor CMI estimator}
Inspired by \citet{Dobrushin1958}, \citet{kozachenko1987sample} introduced a class of differential entropy estimators that can be generalized to estimators of conditional mutual information. This class is based on nearest neighbor statistics as further discussed in \citet{kozachenko1987sample,FrenzelPompe2007}. For a $D_X$-dimensional random variable $X$ the nearest neighbor entropy estimate is defined as 
\begin{align} \label{eq:entropy_knn}
\widehat{H}_{X}  &= \psi(n) +   \frac{1}{n} \sum_{i=1}^n \left[ - \psi(k_{X,i}) + \log(\epsilon_{i}^{D_X}) \right] + \log( V_{D_X})
\end{align}
with the Digamma function as the logarithmic derivative of the Gamma function $\psi(x)=\frac{d }{d x} \ln \Gamma(x)$, sample length $n$, volume element $V$ depending on the chosen metric, i.e., $V_{D_X}=2^{D_X}$ for the maximum metric, $V_{D_X}=\pi^{{D_X}/2}/\Gamma({D_X}/2 + 1)$ for Euclidean metric with Gamma function $\Gamma$. For every sample with index $i$, the integer $k_{X,i}$ is the number of points in the $D_X$-dimensional ball with radius $\epsilon_{i}$. Formula~\eqref{eq:entropy_knn} holds for any $\epsilon_i$ and the corresponding $k_{X,i}$, which will be used in the following definition of a CMI estimator.
Based on this entropy estimator, \citet{Kraskov2004a} derived an estimator for mutual information where the $L_{\infty}$-balls with radius $\epsilon_{i}$ are hypercubes. This estimator was generalized to an estimator for CMI first by \citet{FrenzelPompe2007} and independently by \citet{Vejmelka2008}.
The CMI estimator is obtained by inserting the entropy estimator Eq.~(\ref{eq:entropy_knn}) for the different entropies in the definition of CMI in Eq.~(\ref{eq:def_cmi}). For all entropy terms $H_{XZ},\,H_{YZ},\,H_{Z},\,H_{XYZ}$ in Eq.~(\ref{eq:def_cmi}), we use the maximum norm and choose as the side length $2\epsilon_i$ of the hypercube the distance $\epsilon_i$ to the $k=k_{XYZ,i}$-nearest neighbor in the joint space $\mathcal{X}\otimes \mathcal{Y}\otimes \mathcal{Z}$. The CMI estimate then is
\begin{align} \label{eq:cmi_knn_est}
     &\widehat{I}_{ XY|Z}  \nonumber\\
     &=   \psi (k) + \frac{1}{n} \sum_{i=1}^n \left[ \psi(k_{Z,i}) - \psi(k_{XZ,i}) - \psi(k_{YZ,i}) \right].
\end{align}
The only free parameter $k$ is the number of nearest neighbors in the joint space of $\mathcal{X}\otimes \mathcal{Y}\otimes \mathcal{Z}$ and $k_{{ xz},i}$, $k_{{ yz},i}$ and $k_{{ z},i}$ are computed as follows for every sample point indexed by $i$:
\begin{enumerate}
    \item Determine (here in maximum norm) the distance $\epsilon_i$ to its $k$-th nearest neighbor (excluding the reference point which is not a neighbor of itself) in the joint space $\mathcal{X}\otimes \mathcal{Y}\otimes \mathcal{Z}$.
    \item Count the number of points with distance strictly smaller than $\epsilon_i$ (including the reference point at $i$) in the subspace $\mathcal{X}\otimes \mathcal{Z}$ to get $k_{{ xz},i}$, in the subspace $\mathcal{Y}\otimes \mathcal{Z}$ to get $k_{{ yz},i}$, and in the subspace $\mathcal{Z}$ to get $k_{{ z},i}$.
\end{enumerate}

Similar estimators, but for the more general class of R\'enyi entropies and divergences, were developed in \citet{QingWang2009,Poczos2012b}. 
Estimator~\eqref{eq:cmi_knn_est} uses the approximation that the densities are constant within the epsilon environment. Therefore, the estimator's bias will grow with $k$ since larger $k$ lead to larger $\epsilon$-balls where the assumption of constant density is more likely violated. The variance, on the other hand, is the more important quantity in conditional independence testing and it becomes smaller for larger $k$ because fluctuations in the $\epsilon$-balls average out. 
The decisive advantage of this estimator compared to fixed bandwidth approaches is its \emph{data-adaptiveness} (Fig.~\ref{fig:schematic}B).

The Kozachenko-Leonenko estimator is asymptotically unbiased and consistent \citep{kozachenko1987sample,Leonenko2008a}.
Unfortunately, at present there are no results, neither exact nor asymptotically, on the distribution of the estimator as needed to derive analytical significance bounds. In \citet{Goria2005}, some numerical experiments indicate that for many distributions of $X,\,Y$ the asymptotic distribution of MI is Gaussian. But the important finite size dependence on the dimensions $D_X,\,D_Y,\,D_Z$, the sample length $n$ and the parameter $k$ are unknown.

Some notes on the implementation:
Before estimating CMI, we rank-transform the samples individually in each dimension: Firstly, to avoid points with equal distance, small amplitude random noise is added to break ties. Then, for all $n$ values $x_1, \ldots, x_n$, we replace $x_i$ with the transformed value $r$, where $r$ is defined such that $x_i$ is the $r$th largest among all $x$ values.
The main computational cost comes from searching nearest neighbors in the high dimensional subspaces which we  speed up using \emph{KD-tree} neighbor search \citep{Maneewongvatana1999}. Hence, the computational complexity will typically scale less than quadratically with the sample size. Kernel methods, on the other hand, typically scale worse than quadratically in sample size if they are not based on Kernel approximations such as via random Fourier features \citep{Strobl2017}. Further, the CMI estimator scales roughly linearly in $k$ and $D$, the total dimension of $X,Y,Z$.

\subsection{Nearest-neighbor permutation test}
\begin{figure}[ht]
\centering
\includegraphics[width=0.8\linewidth]{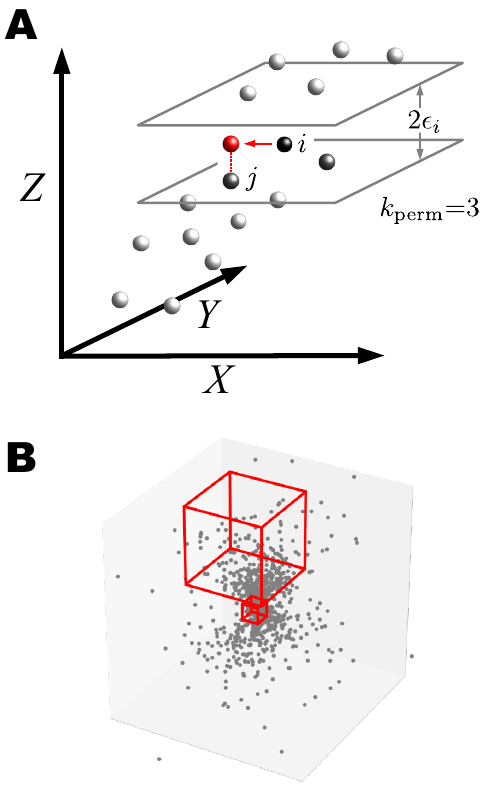}
\caption{(A) Schematic of local permutation scheme. Each sample point $i$'s $x$-value is mapped randomly to one of its $k_{\rm perm}$-nearest neighbors in subspace $\mathcal{Z}$. The hypercubes with length scale $\epsilon_i$ locally adapt to the density making this scheme more data efficient than fixed bandwidth techniques. By keeping track of already `used' indices $j$, we approximately achieve a random draw \emph{without} replacement, see Algorithm~\ref{alg:permutation}.
(B) The CMIknn estimator and the local permutation test are data-adaptive: The hypercubes around each sample point are smaller where more samples are available.}
\label{fig:schematic}
\end{figure}
\begin{algorithm}[h]
\caption{Algorithm to generate a nearest-neighbor permutation $\pi(\cdot)$ of $\{0,1,\ldots,n\}$.}
\begin{algorithmic}[1]
\State Denote by $d^{k_{\rm perm}}_i$ the distance of sample point $z_i$ to its $k_{\rm perm}$-nearest neighbor (including $i$ itself, i.e., $d^{k_{\rm perm}=1}_i=0$)
\State Compute list of nearest neighbors for each sample point: $\mathcal{N}_i=\{l \in \{0,\ldots,n\}: \|z_l - z_i\| \leq d^{k_{\rm perm}}_i \}$ with KD-tree algorithm in maximum norm of subspace $Z$
\State Shuffle $\mathcal{N}_i$ for each $i$
\State Initialize empty list $\mathcal{U}=\{\}$ of used indices
\ForAll{$i\in$ random permutation of $\{1,\ldots,n\}$ }
        \State $j=\mathcal{N}_i(0)$
        \State $m = 0$
        \While{$j \in \mathcal{U}$ and $m < k_{\rm perm} -1$}
            \State $m=m+1$
            \State $j=\mathcal{N}_i(m)$
        \EndWhile
        \State $\pi(i) = j$
        \State Add $j$ to $\mathcal{U}$
\EndFor
\State \Return $\{\pi(1),\ldots,\pi(n)\}$
\end{algorithmic}
\label{alg:permutation}
\end{algorithm}

Since no theory on finite sample behavior of the CMI estimator is available, we resort to a permutation-based generation of the distribution under $H_0$.

Typically in CMI-based independence testing, CMI-surrogates to simulate independence are generated by randomly permuting \emph{all} values in $X$. The problem is, that this approach not only destroys the dependence between $X$ and $Y$, as desired, but also destroys all dependence between $X$ and $Z$. Hence, this approach does not actually test $X ~\ci ~Y~ |~ Z$. In order to preserve the dependence between  $X$ and $Z$, we propose a local permutation test utilizing nearest-neighbor search. To avoid confusion, we denote the CMI-estimation parameter as $k_{\rm CMI}$ and the permutation-parameter as $k_{\rm perm}$. 

As illustrated in Fig.~\ref{fig:schematic}, we first identify the $k_{\rm perm}$-nearest neighbors around each sample point $i$ (here including the point itself) in the subspace of $Z$ using the maximum norm. With Algorithm~\ref{alg:permutation} we generate a permutation mapping $\pi: \{1,\ldots,n\} \to \{\pi(1),\ldots,\pi(n)\}$ which tries to achieve draws \emph{without} replacement. Since this is not always possible, some values might occur more than once, i.e., they were drawn \emph{with} replacement as in a bootstrap. 
In \citet{Paparoditis2000} a bootstrap scheme that \emph{always} draws with replacement is described which is used for the CDC independence test. The difference to our scheme is that we try to avoid tied samples as much as possible to preserve the conditional marginals.

The permutation test is then as follows: 
\begin{enumerate}
    \item Generate a random permutation $x^*_b=\{x_{\pi_b(1)}, \ldots, x_{\pi_b(n)}\}$ with Algorithm~\ref{alg:permutation}
    \item Compute CMI $\widehat{I}(x^*_b;y|z)$ via Eq.~\eqref{eq:cmi_knn_est}
    \item Repeat steps (1) and (2) $B$ times, sort the estimates $\widehat{I}_b$ from the null  and obtain $p$-value by 
        \begin{align}
            p = \frac{1}{B} \sum_{b=1}^B  \mathbbm{1}_{\widehat{I}_b\geq \widehat{I}(x;y|z)} \,,
        \end{align}
        where $\mathbbm{1}$ denotes the indicator function and $\widehat{I}(x;y|z)$ is the CMI estimate of the original data.
\end{enumerate}
The CMI estimator holds for arbitrary dimensions of all arguments $X,Y,Z$ and also the local permutation scheme can be used to jointly permute all of $X$'s dimensions. In the following numerical experiments, we focus on the case of univariate $X$ and $Y$ and uni- or multivariate $Z$.

\section{Experiments}

\subsection{Choosing $k_{\rm CMI}$ and $k_{\rm perm}$}
\begin{figure}[ht]
\centering
\includegraphics[width=1.\linewidth]{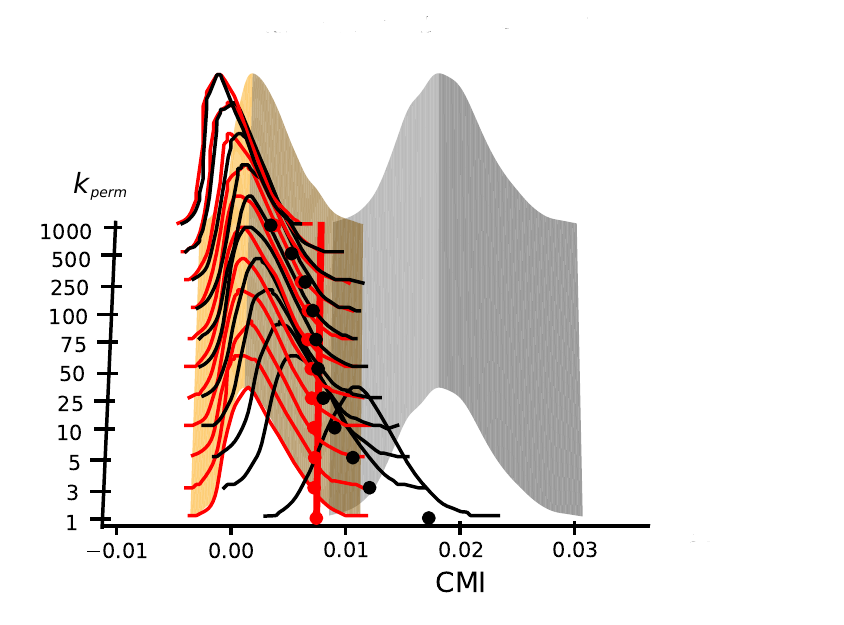}%
\caption{Illustrative simulation of multivariate Gaussian to demonstrate the effect of the nearest-neighbor permutation parameter $k_{\rm perm}$. The true null distribution of CMI is depicted as the orange surface with the 5\% quantile marked by a red straight line, and the true distribution under dependence is drawn as a grey surface (both are constant for all $k_{\rm perm}$). The red and black distributions and markers give the permuted null distributions and their 5\% quantiles for different $k_{\rm perm}$ for the independent (red) and dependent (black) case, respectively. Here the sample size is $n=1000$ such that $k_{\rm perm}=1000$ corresponds to a full non-local permutation.}
\label{fig:dists}
\end{figure}
\begin{figure*}[ht]
\newcommand{\lw}{.8}
\centering
\includegraphics[width=\lw\linewidth]{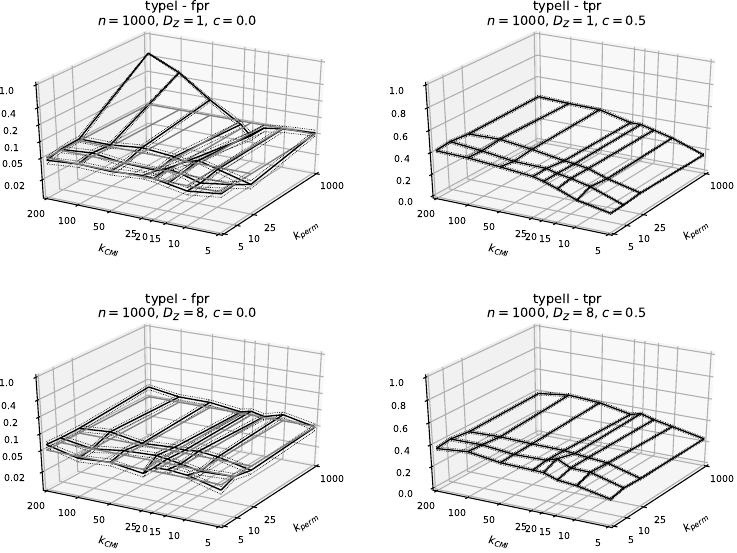}
\caption{Numerical experiments with post-nonlinear noise model \citep{Zhang2012,Strobl2017}. The sample size is $n=1000$ and $1000$ realizations were generated to evaluate false positives (FPR) and true positives (TPR) for $c=0.5$  at the 5\% significance level. Shown are FPR and TPR for $D_Z=1$ (top panels) and $D_Z=8$ (bottom panels).}
\label{fig:knn_neigh}
\end{figure*}
\begin{figure}[ht]
\newcommand{\lw}{1.}
\centering
\includegraphics[width=\lw\linewidth]{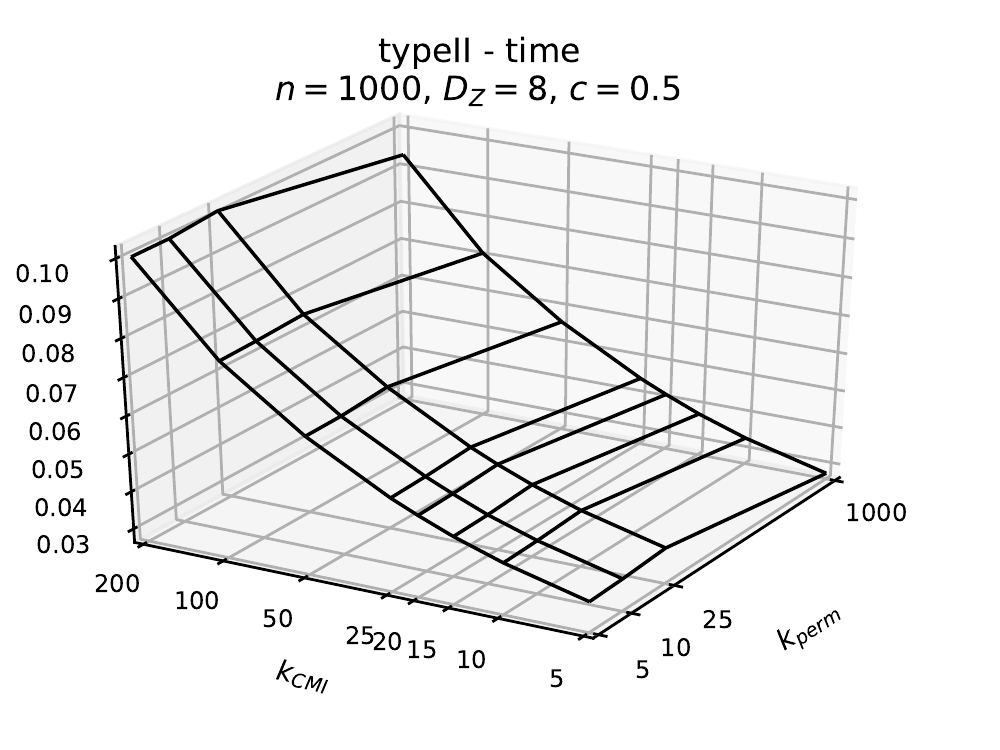}
\caption{Runtime per estimate [in s] for the same setup as in the lower panels of Fig.~\ref{fig:knn_neigh}. For $k_{\rm perm}=n$ a computationally cheaper full permutation scheme was used.}
\label{fig:knn_neigh_time}
\end{figure}

Our approach has two free parameters $k_{\rm CMI}$ and $k_{\rm perm}$. The following numerical experiments indicate that restricting $k_{\rm perm}$ to only very few nearest neighbors already suffices to reliably simulate the null distribution in most cases while for $k_{\rm CMI}$ we derive a rule-of-thumb based on the sample size $n$.

Figure~\ref{fig:dists} illustrates the effect of different $k_{\rm perm}$. If $k_{\rm perm}$ is too large or even  $k_{\rm perm}=n$, i.e., a full non-local permutation, the permuted distribution under independence (red) is negatively biased. As illustrated by the red markers, this would lead to an increase of false positives (type-I error). On the other hand, for the dependent case, if $k_{\rm perm}=1..3$, the permuted distribution is positively biased yielding lower power (type-II errors). For a range of optimal values of $k_{\rm perm}$, the permuted distribution perfectly simulates the true null distribution.

To evaluate the effect of $k_{\rm CMI}$ and $k_{\rm perm}$ numerically, we followed the post-nonlinear noise model described in \citet{Zhang2012,Strobl2017} given by  $X = g_X (\epsilon_X + \frac{1}{D_Z}\sum^{D_Z}_i Z_i)$, $Y = g_Y (\epsilon_Y + \frac{1}{D_Z}\sum^{D_Z}_i Z_i )$, where $Z_i, \epsilon_X , \epsilon_Y$ have jointly independent standard Gaussian distributions, and $g_X , g_Y$ denote smooth functions uniformly chosen from ${(\cdot), (\cdot)^2 , (\cdot)^3 , \tanh(\cdot), \exp(||\cdot||^2)}$. Thus, we have $X ~\ci ~Y~ |~ Z=(Z_1,Z_2,\ldots)$ in any case. To simulate dependent $X$ and $Y$, we used $X = g_X (c\epsilon_b +\epsilon_X)$, $Y = g_Y (c\epsilon_b +\epsilon_Y )$ for $c>0$ and identical Gaussian noise $\epsilon_b$ and keep $Z$ independent of $X$ and $Y$.

\begin{figure*}[ht]
\newcommand{\lw}{.98}
\centering
\includegraphics[width=\lw\linewidth]{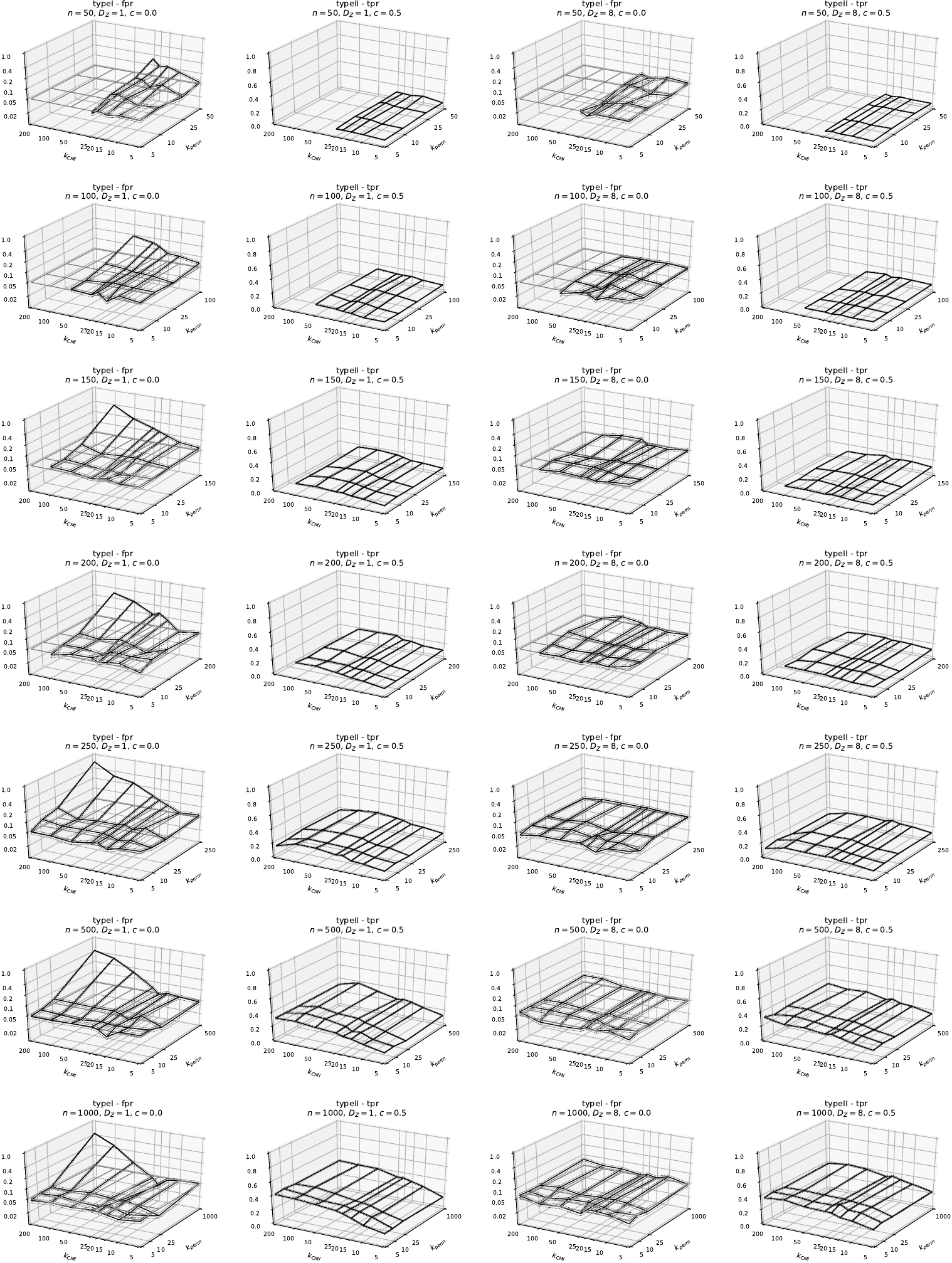}
\caption{Same as in Fig.~\ref{fig:knn_neigh}, but for more sample sizes from $n=50$ (top) to $n=1000$ (bottom).}
\label{fig:knn_neigh_all}
\end{figure*}

In Fig.~\ref{fig:knn_neigh}, we show results for sample size $n=1000$. The null distribution was generated with $B=1000$ surrogates in all experiments.
The results demonstrate that a value  $k_{\rm perm}\approx 5..10$ yields well-calibrated tests while not affecting power much. This holds for a wide range of sample sizes as shown in Fig.~\ref{fig:knn_neigh_all}. 

Larger $k_{\rm CMI}$ yield more power and even for $k_{\rm CMI}\approx n/2$ the tests are still well calibrated. But power peaks at some value of $k_{\rm CMI}$ and slowly decreases for too large values. Still, the dependency of power on $k_{\rm CMI}$ is relatively robust and we suggest a rule-of-thumb of $k_{\rm CMI}\approx 0.1..0.2 n$. Note that, as shown in Fig.~\ref{fig:knn_neigh_time}, runtime increases linearly with $k_{\rm CMI}$  while $k_{\rm perm}$ does not impact runtime much. Here we depict the runtime per CMI estimate assuming that the permutation scheme is (embarrassingly) parallelized.

\subsection{Comparison with kernel measures}
\begin{figure*}[ht]
\newcommand{\lw}{1.}
\centering
\includegraphics[width=\lw\linewidth]{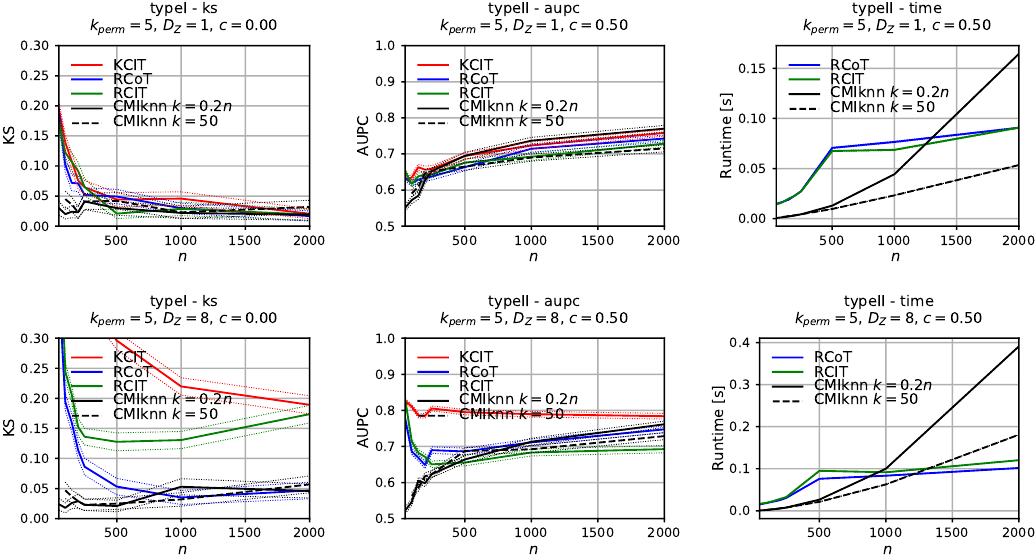}
\caption{Numerical experiments with post-nonlinear noise model and similar setup as in \citet{Strobl2017}. Shown are KS (left column), AUPC (center column), and runtime (right column) for a sample size experiment with $D_Z=1$ (top row) and $D_Z=8$ (center row). In all experiments we set $k_{\rm perm}=5$ and depict $k_{\rm CMI}=0.2 n$ and $k_{\rm CMI}=50$. 
Here we show results for the default $n_{\text{ff}}=25$ fourier features for RCIT and RCoT, but much more are needed to resolve less smooth densities (Figs.~\ref{fig:sine},\ref{fig:mult}). In Fig.~\ref{fig:strobl_paras} we show more parameter studies for RCIT and RCoT.}
\label{fig:strobl_ksaupc}
\end{figure*}
\begin{figure*}[ht]
\newcommand{\lw}{1.}
\centering
\includegraphics[width=0.7\linewidth]{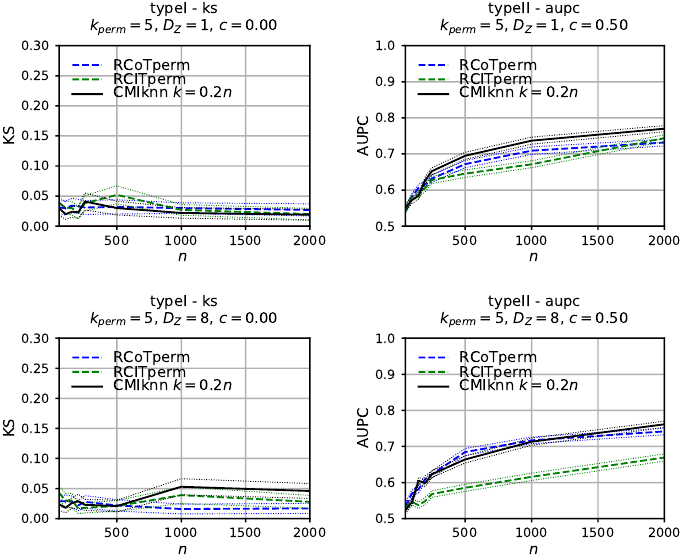}
\caption{Numerical experiments as before for the kernel measures combined with the proposed nearest-neighbor permutation test with $k_{\rm perm}=5$. 
}
\label{fig:strobl_ksaupc_shuffle}
\end{figure*}
\begin{figure*}[ht]
\newcommand{\lw}{1.}
\centering
\includegraphics[width=\lw\linewidth]{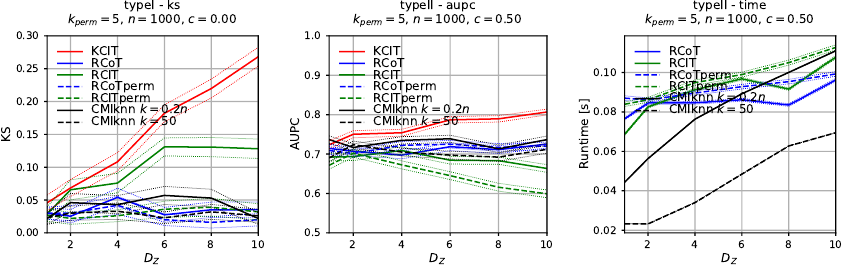}
\caption{Numerical experiments as before for different condition dimensions $D_Z$ with fixed $n=1000$.}
\label{fig:strobl_ksaupc_dims}
\end{figure*}
\begin{figure*}[ht]
\newcommand{\lw}{1.}
\centering
\includegraphics[width=\lw\linewidth]{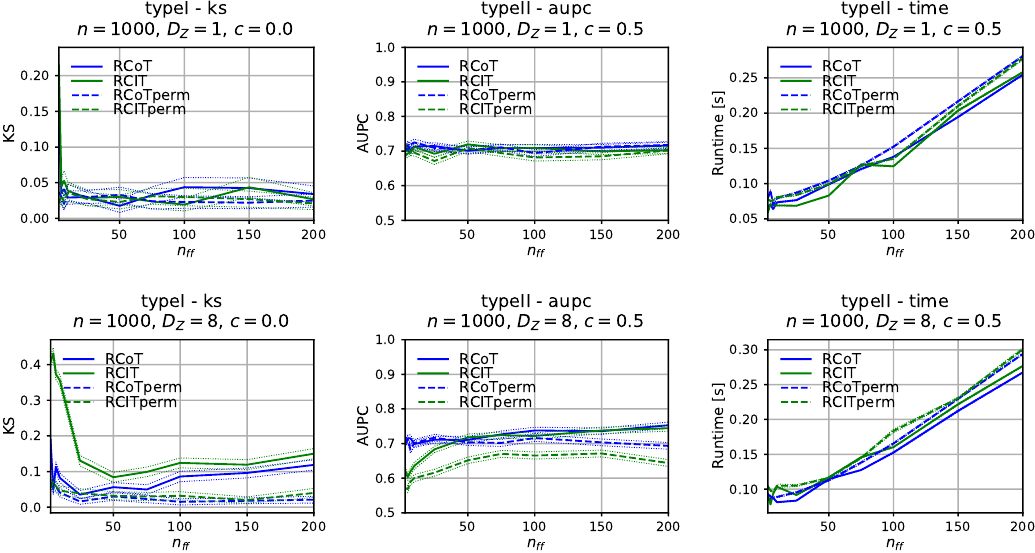}
\caption{Choice of number of fourier features ($n_{\text{ff}}$) for random fourier-feature based kernel-measures. Experiments based on post-nonlinear noise model and similar setup as in \citet{Strobl2017}. Shown are KS (left column), AUPC (center column), and runtime (right column) for a sample size experiment with $D_Z=1$ (top row) and $D_Z=8$ (center row). $n_{\text{ff}}$ corresponds to the number of features in subspace $Z$, the number of fourier features in subspaces $X$ and $Y$ is fixed to $5$ as implemented in \texttt{https://github.com/ericstrobl/RCIT}. Solid lines mark RCIT and RCoT tests based on analytically approximating the null distribution while dashed lines are based on the nearest-neighbor local permutation scheme introduced in this work. While for $D_Z=1$ $n_{\text{ff}}>10$ yields similar results, for $D_Z=8$ the KS metric is more sensitive to the choice of $n_{\text{ff}}$, at least for the analytical RCIT and RCoT versions.
The runtime of RCIT and RCoT scales roughly quadratically in the number of fourier features. 
}
\label{fig:strobl_paras}
\end{figure*}
\begin{figure*}[ht]
\newcommand{\lw}{.7}
\centering
\includegraphics[width=.5\linewidth]{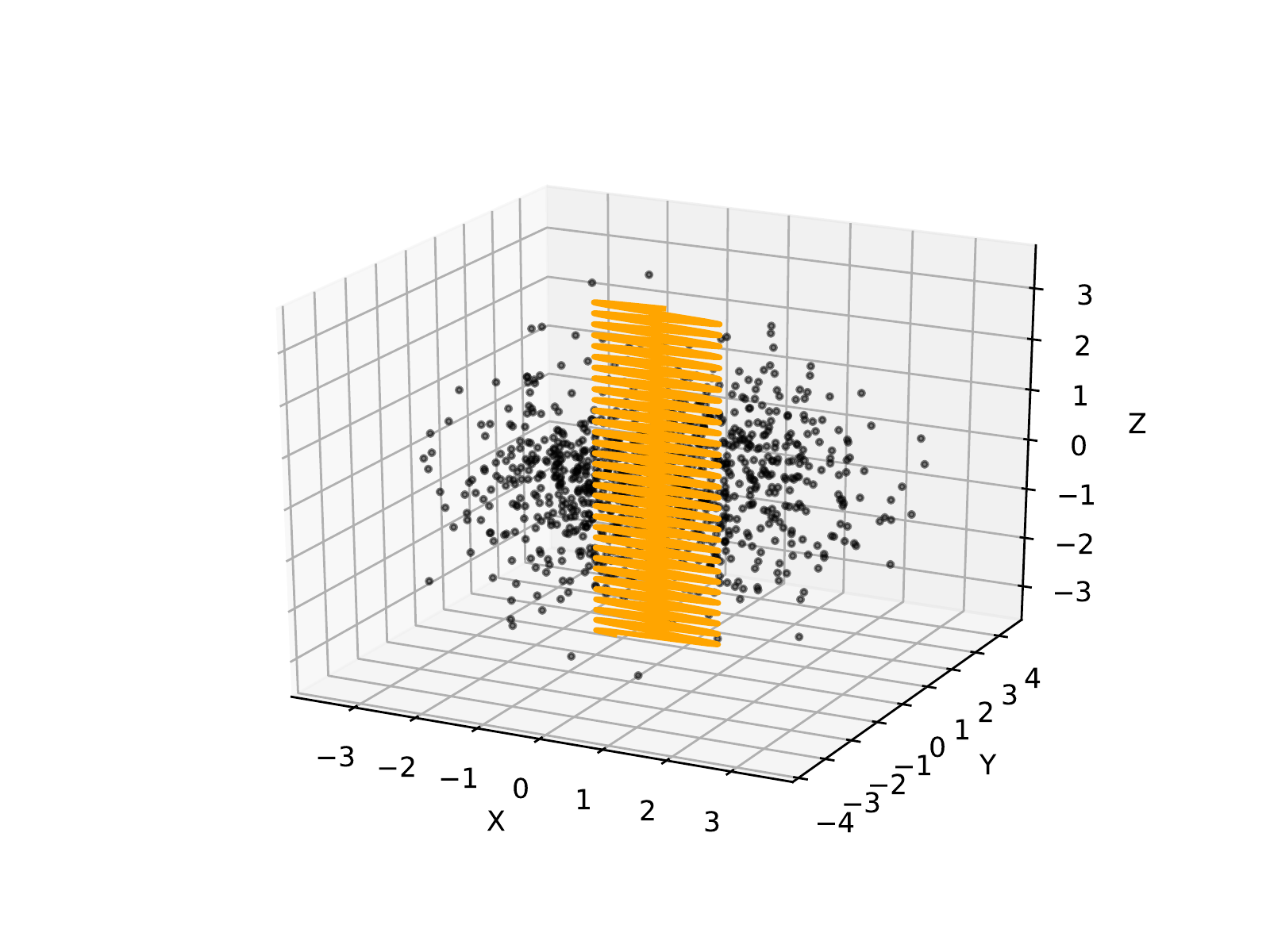}
\includegraphics[width=\lw\linewidth]{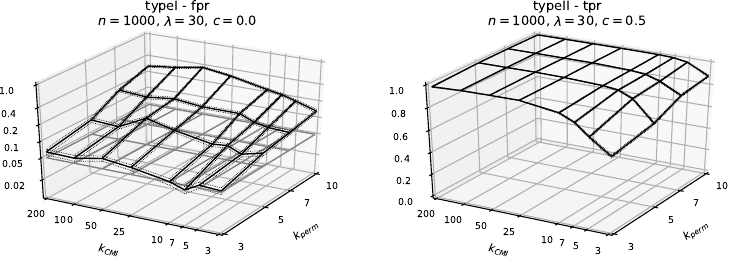}
\includegraphics[width=\lw\linewidth]{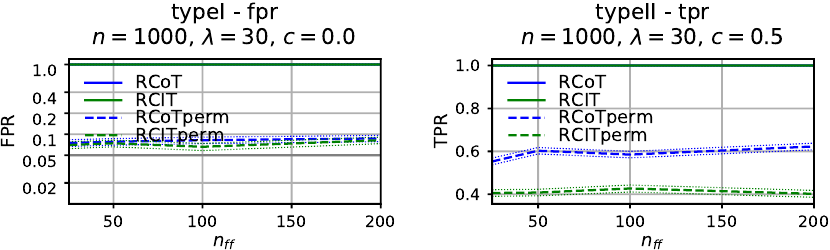}
\caption{Example of sinusoidal dependence $X=\sin(\lambda Z) + \epsilon_X$ and $Y=\sin(\lambda Z) + \epsilon_Y$ leading to strongly oscillatory structure (top panel for $\lambda=30$). The second row depicts FPR and TPR for CMIknn and the bottom row for RCIT and RCoT for different numbers of random fourier features $n_{\text{ff}}$. Here the analytical versions of RCIT and RCoT (solid lines) do not work at all (FPR equal to 1). The permutation versions of RCIT and RCoT (dashed lines) use $k_{\rm perm}=5$.}
\label{fig:sine}
\end{figure*}
\begin{figure*}[ht]
\newcommand{\lw}{.7}
\centering
\includegraphics[width=.5\linewidth]{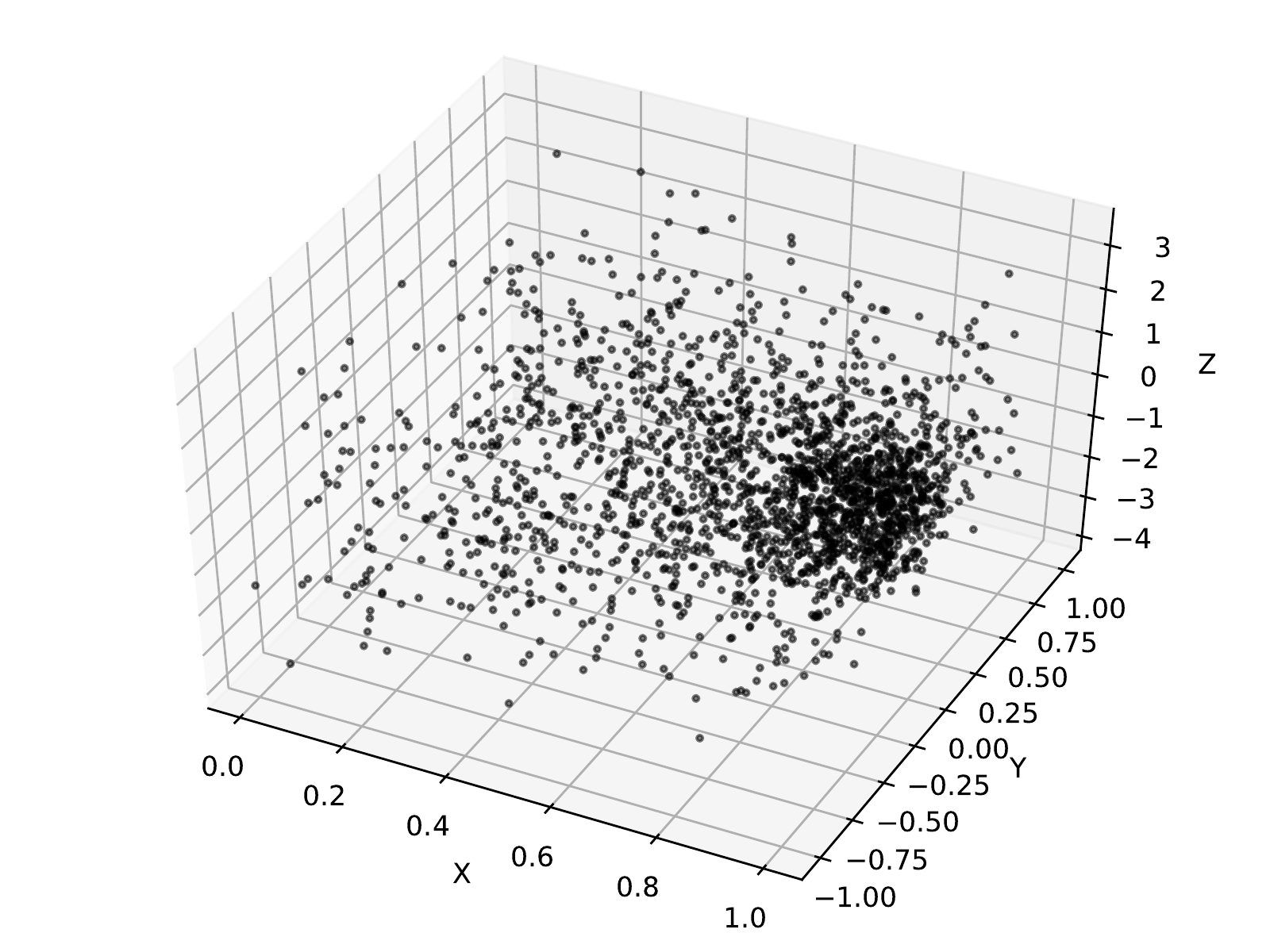}
\includegraphics[width=\lw\linewidth]{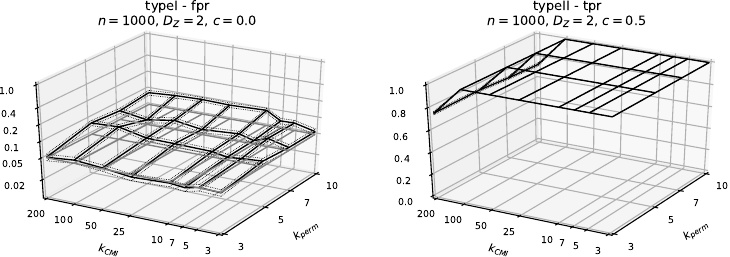}
\includegraphics[width=\lw\linewidth]{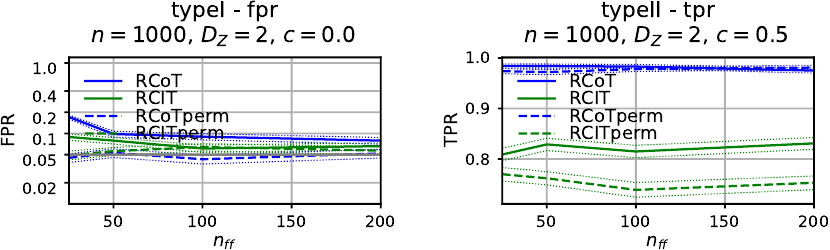}
\caption{Example of multiplicative dependence of $X$ and $Y$ on $Z$ leading to strongly nonlinear structure (top panel). 
Here the nearest-neighbor scheme of CMIknn can better adapt to the very localized density $D_Z=2$ (center panel) with $k_{\rm perm}<7$ while RCIT and RCoT (bottom panel) do not well control false positives even if we resolve smaller scales better using a larger number of Fourier features $n_{\text{ff}}$. The permutation-versions are better calibrated, but have less power, especially RCIT.
}
\label{fig:mult}
\end{figure*}

In Fig.~\ref{fig:strobl_ksaupc} we show results comparing our CMI test (CMIknn) to KCIT and the two random-fourier-based approximations RCIT and RCoT introduced in \citet{Strobl2017}.  As a metric for type-I errors, as in \citet{Strobl2017} we evaluated the Kolmogorov-Smirnov (KS) statistic to quantify how uniform the distribution of $p$-values is. For type-II errors we measure the area under the power curve (AUPC). All metrics were evaluated from $1000$ realizations and error bars give the boostrapped standard errors. 
We show results for CMIknn for $k_{\rm perm}=5$ with $B=1000$ permutation surrogates and using the rule-of-thumb $k_{\rm CMI}=0.2n$ as well as a fixed $k_{\rm CMI}=50$.

Figure~\ref{fig:strobl_ksaupc} demonstrates that CMIknn is better calibrated with the lowest KS-values for almost all sample sizes tested. KCIT and RCIT are especially badly calibrated for smaller sample sizes or higher dimensions $D_Z$ and RCoT better approximates the null distribution only for $n\geq 500$ for $D_Z=1$ and for $n\geq 1000$ for $D_Z=8$. Note that this is also expected \citep{Strobl2017} since the analytical approximation of the null distribution for RCIT and RCoT requires large sample sizes. The power as measured by AUPC is, thus only comparable for $n>500$ for $D_Z=1$ and CMIknn has the highest power throughout if $k_{\rm CMI}$ is scaled with the sample size. Also for fixed $k_{\rm CMI}=50$ the power of CMIknn is competitive. Also for $D_Z=8$ and $n\geq 1000$ CMIknn has slightly higher power than RCoT and RCIT. 

If the computationally expensive permutation scheme of CMIknn is (embarrassingly) parallelized, the CMIknn test is  faster than RCIT or RCoT for not too large sample sizes due to efficient KD-tree nearest-neighbor search procedures \citep{Maneewongvatana1999}, especially for smaller $k_{\rm CMI}$. KCI is not shown here because it is orders of magnitude slower. A major computational burden of RCIT and RCoT is the kernel bandwith computation via the median Euclidean distance heuristic. In \texttt{https://github.com/ericstrobl/RCIT} the median is computed from the first $500$ samples only, leading to the ``kink'' in the runtime for RCIT and RCoT. 
The runtime of RCIT and RCoT depends quadratically on the number of random Fourier features used (here the default of $25$ for subspace $\mathcal{Z}$ and $5$ for subspaces $\mathcal{X}$ and $\mathcal{Y}$ was used), for more results see Fig.~\ref{fig:strobl_paras}. CMIknn's runtime increases more sharply with sample size.

Our results indicate that the analytical approximations of the null distribution utilized in RCIT and RCoT do not work well for small sample sizes below $n\approx 1000$. In Fig.~\ref{fig:strobl_ksaupc_shuffle} we explore the option to combine the kernel statistics with our nearest-neighbor permutation test. While then RCIT and RCoT loose their computational advantage, the tests are now well-calibrated. Their power is still mostly lower than that of CMIknn, especially for RCIT.

In Fig.~\ref{fig:strobl_ksaupc_dims} we explore more cardinalities of the conditioning set. KCI and RCIT are not well-calibrated for higher dimensions and also the permutation-version of RCIT quickly looses power for higher dimensions. The power of CMIknn and RCoT is rather insensitive to the dimensionality. Note, however, that in the numerical experiments of \citet{Strobl2017} the conditioning variables $Z$ for evaluating power are independent of $X$ and $Y$. Other experimental setups might induce a dependence of power on $D_Z$. CMIknn's runtime starts lower, but increases more sharply with $D_Z$ than RCIT and RCoT.

Until now we considered rather smooth dependencies of $X$ and $Y$ on the conditioning variables. In Figs.~\ref{fig:sine},\ref{fig:mult} we consider more nonlinear relationships. For an extremely oscillatory sinusoidal dependency like $X=\sin(\lambda Z) + \epsilon_X$ and $Y=\sin(\lambda Z) + \epsilon_Y$ ($c\epsilon_b$ added for the dependent case), shown in Fig.~\ref{fig:sine}, $k_{\rm perm}$ needs to be set to a very small value in order to control false positives. Here the analytical versions of RCIT and RCoT do not work at all and the permutation-based versions have much lower power than CMIknn.

In Fig.~\ref{fig:mult} we consider a multiplicative noise case with the model $X = g_X (0.1 \epsilon'_X +\epsilon_X \frac{1}{D_Z}\sum^{D_Z}_i Z_i)$, $Y = g_Y (0.1 \epsilon'_Y + \epsilon_Y\frac{1}{D_Z}\sum^{D_Z}_i Z_i )$ with all variables as before and $\epsilon'_{X,Y}$ another independent Gaussian noise term. For the dependent case we used $X = g_X (c\epsilon_b \epsilon_X)$, $Y = g_Y (c\epsilon_b \epsilon_Y )$ for $c>0$ and identical Gaussian noise $\epsilon_b$ and keep $Z$ independent of $X$ and $Y$.
Even though the density is highly localized in this case, CMIknn is still well calibrated for $k_{\rm perm}\approx 5$. On the other hand, RCoT cannot control false positives even if we vary the number of Fourier features to much higher values (which takes much longer). RCIT is slightly better suited here, but only combined with the local permutation test both become better calibrated. CMIknn has higher power than both permutation-based kernel tests in this example.

\subsection{Comparison with conditional distance correlation}
\begin{table*}[ht]
\centering
\caption{Results from \citet{Wang2015} together with results from RCoT and our CMI test. The experiments are described in \citet{Wang2015}. Examples 1--4 correspond to conditional independence showing false positives and Examples 5--8 to dependent cases showing true positives at the 5\% significance level. CMIknn was run with $k_{\rm CMI}=0.2n$  and $k_{\rm perm}=5,10$. The numbers $50..250$ denote the sample size.}
\label{tab:wang}
\begin{tabular}{@{}l|llllll|lllll@{}}
\toprule
              & \multicolumn{5}{c}{Example 1}         &  & \multicolumn{5}{c}{Example 2}         \\
Test          & 50    & 100   & 150   & 200   & 250   &  & 50    & 100   & 150   & 200   & 250   \\
\hline
CDIT          & 0.035 & 0.034 & 0.05  & 0.057 & 0.048 &  & 0.046 & 0.053 & 0.055 & 0.048 & 0.058 \\
CI.test       & 0.041 & 0.051 & 0.037 & 0.054 & 0.041 &  & 0.062 & 0.046 & 0.044 & 0.045 & 0.039 \\
KCI.test      & 0.039 & 0.043 & 0.041 & 0.04  & 0.046 &  & 0.035 & 0.004 & 0.037 & 0.047 & 0.05  \\
Rule-of-thumb & 0.017 & 0.027 & 0.028 & 0.033 & 0.033 &  & 0.034 & 0.052 & 0.044 & 0.042 & 0.045 \\
RCoT & 0.074  & 0.059  & 0.055  & 0.043  & 0.050  &  & 0.056  & 0.056  & 0.069  & 0.055  & 0.073 \\
CMIknn ($k_{\rm perm}=5$) & 0.064  & 0.055  & 0.050  & 0.053  & 0.045  &  & 0.076  & 0.060  & 0.074  & 0.061  & 0.065 \\
CMIknn ($k_{\rm perm}=10$) & 0.058  & 0.061  & 0.057  & 0.058  & 0.046  &  & 0.075  & 0.066  & 0.053  & 0.057  & 0.071 \\
             &       &       &       &       &       &  &       &       &       &       &       \\
              & \multicolumn{5}{c}{Example 3}         &  & \multicolumn{5}{c}{Example 4}         \\
Test          & 50    & 100   & 150   & 200   & 250   &  & 50    & 100   & 150   & 200   & 250   \\
\hline
CDIT          & 0.035 & 0.048 & 0.055 & 0.053 & 0.043 &  & 0.049 & 0.054 & 0.051 & 0.058 & 0.053 \\
CI.test       & 0.222 & 0.363 & 0.482 & 0.603 & 0.677 &  & 0.043 & 0.064 & 0.066 & 0.05  & 0.053 \\
KCI.test      & 0.058 & 0.047 & 0.057 & 0.061 & 0.054 &  & 0.037 & 0.035 & 0.058 & 0.039 & 0.049 \\
Rule-of-thumb & 0.019 & 0.038 & 0.032 & 0.039 & 0.039 &  & 0.037 & 0.04  & 0.055 & 0.059 & 0.053 \\
RCoT & 0.074  & 0.047  & 0.046  & 0.053  & 0.054  &  & 0.115  & 0.072  & 0.066  & 0.061  & 0.053 \\
CMIknn ($k_{\rm perm}=5$) & 0.044  & 0.043  & 0.046  & 0.046  & 0.054  &  & 0.084  & 0.071  & 0.067  & 0.079  & 0.070 \\
CMIknn ($k_{\rm perm}=10$) & 0.063  & 0.065  & 0.061  & 0.076  & 0.067  &  & 0.101  & 0.113  & 0.106  & 0.098  & 0.084 \\
              &       &       &       &       &       &  &       &       &       &       &       \\
              & \multicolumn{5}{c}{Example 5}         &  & \multicolumn{5}{c}{Example 6}         \\
Test          & 50    & 100   & 150   & 200   & 250   &  & 50    & 100   & 150   & 200   & 250   \\
\hline
CDIT          & 0.898 & 0.993 & 1     & 1     & 1     &  & 0.752 & 0.995 & 1     & 1     & 1     \\
CI.test       & 0.978 & 1     & 1     & 1     & 1     &  & 0.468 & 0.434 & 0.467 & 0.476 & 0.474 \\
KCI.test      & 0.158 & 0.481 & 0.557 & 0.602 & 0.742 &  & 0.296 & 0.862 & 0.995 & 1     & 1     \\
Rule-of-thumb & 0.368 & 0.793 & 0.927 & 0.983 & 0.994 &  & 1     & 1     & 1     & 1     & 1     \\
RCoT & 0.817  & 0.986  & 0.998  & 1  & 1  &  & 0.301  & 0.533  & 0.679  & 0.807  & 0.860 \\
CMIknn ($k_{\rm perm}=5$) & 0.782  & 0.981  & 0.998  & 1  & 1  &  & 0.806  & 0.997  & 0.999  & 1  & 1 \\
CMIknn ($k_{\rm perm}=10$) & 0.855  & 0.995  & 1  & 1  & 1  &  & 0.805  & 0.995  & 1  & 1  & 1 \\
              &       &       &       &       &       &  &       &       &       &       &       \\
              & \multicolumn{5}{c}{Example 7}         &  & \multicolumn{5}{c}{Example 8}         \\
Test          & 50    & 100   & 150   & 200   & 250   &  & 50    & 100   & 150   & 200   & 250   \\
\hline
CDIT          & 0.918 & 0.998 & 1     & 1     & 1     &  & 0.361 & 0.731 & 0.949 & 0.977 & 0.994 \\
CI.test       & 0.953 & 0.984 & 0.983 & 0.995 & 0.987 &  & 0.456 & 0.476 & 0.464 & 0.461 & 0.485 \\
KCI.test      & 0.574 & 0.947 & 0.998 & 1     & 1     &  & 0.089 & 0.401 & 0.685 & 1     & 1     \\
Rule-of-thumb & 0.073 & 0.302 & 0.385 & 0.514 & 0.515 &  & 0.043 & 0.233 & 0.551 & 0.851 & 0.972 \\
RCoT & 0.594  & 0.880  & 0.962  & 0.985  & 0.991  &  & 0.275  & 0.392  & 0.470  & 0.624  & 0.654 \\
CMIknn ($k_{\rm perm}=5$) & 0.753  & 0.963  & 0.992  & 0.997  & 1  &  & 0.302  & 0.644  & 0.804  & 0.916  & 0.958 \\
CMIknn ($k_{\rm perm}=10$) & 0.798  & 0.976  & 0.999  & 0.999  & 0.999  &  & 0.323  & 0.680  & 0.832  & 0.920  & 0.971 \\
\bottomrule
\end{tabular}
\end{table*}

In Tab.~\ref{tab:wang} we repeat the results from \citet{Wang2015} proposing the CDC test together with results from RCoT and our CMI test. The experiments are described in \citet{Wang2015}. Examples 1--4 correspond to conditional independence and Examples 5--8 to dependent cases. CMIknn has well-calibrated tests except for Example~4 (as well as Example~8) which is based on discrete Bernoulli random variables while the CMI test is designed for continuous variables.
For Examples 5--8 CMIknn has competitive power compared to CDC and outperforms KCIT in all and RCoT in all but Example~5 where they reach the same performance. Note that the CDC test also is based on a computationally expensive local permutation scheme since the asymptotics break down for small sample sizes. 

\section{Real data application}
\begin{figure}[ht]
\newcommand{\lw}{.97}
\centering
\includegraphics[width=\lw\linewidth]{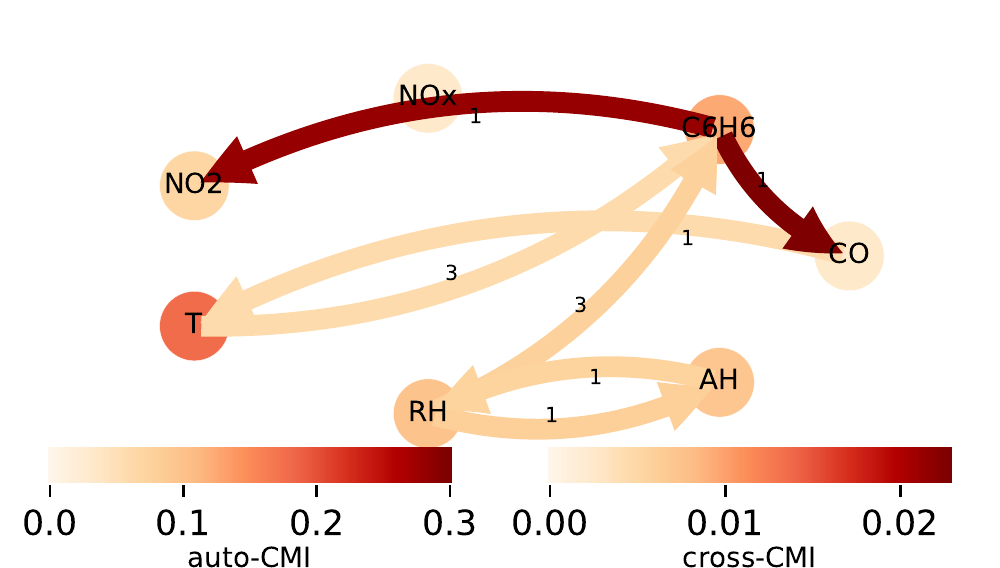}
\caption{Causal discovery in time series of air pollutants and various weather variables. The node color gives the strength of auto-CMI and the edge color the cross-CMI with the link labels denoting the time lag in $hours$. }
\label{fig:real}
\end{figure}
We apply CMIknn in a time series version of the PC causal discovery algorithm \citep{Runge2017} to investigate dependencies between hourly averaged concentrations for carbon monoxide (CO), benzene (C6H6), total nitrogen oxides (NOx), nitrogen dioxide (NO2), as well as temperature (T), relative humidity (RH) and absolute humidity (AH) taken from \citet{DeVito2008}\footnote{\texttt{http://archive.ics.uci.edu/ml/datasets/Air+Quality}}. The time series were detrended using a Gaussian kernel smoother with bandwidth $\sigma=1440~hours$ and we limited the analysis to the first three months of the dataset ($2160$ samples). After accounting for missing values we obtain an effective sample size of $n=1102$. As in our numerical experiments, we used the CMIknn parameters $k_{\rm CMI}=200$ and $k_{\rm perm}=5$ with $B=1000$ permutation surrogates. The causal discovery algorithm was run including lags from $\tau=1$ up to $\tau_{\max}=3~hours$. 
The resulting graph at a 10\% FDR-level shown in Fig.~\ref{fig:real} indicates that temperature and relative humidity influence Benzene which in turn affects NO2 and CO concentrations.

\section{Conclusion}
We presented a novel fully non-parametric conditional independence test based on a nearest neighbor estimator of conditional mutual information. Its main advantage lies in the ability to adapt to highly localized densities due to nonlinear dependencies even in higher dimensions. This feature results in well-calibrated tests with reliable false positive rates. We tested setups for sample sizes $n=50$ to $n=2000$ and dimensions of the conditional set of $D_Z=1..10$. The power of CMIknn is comparable or higher than advanced kernel based tests such as KCIT or its faster random Fourier feature versions RCIT and RCoT, which, however, are not well-calibrated in the smaller sample limit. Combining our local permutation scheme with kernel tests leads to better calibration, but power is still lower than CMIknn.
CMIknn has a shorter runtime for not too large sample sizes since efficient nearest-neighbor search schemes can be utilized, but its runtime increases more sharply with sample size and dimensionality than the fourier-feature bases kernel tests. Here approximate nearest-neighbor techniques could speed up computations. 
The permutation scheme leads to a higher computational load which, however, can be easily parallelized. Nevertheless, more theoretical research is desirable to obtain approximate analytics for the null distribution in the large sample limit. For small sample sizes below $n\approx 1000$ we find that a permutation-approach is inevitable also for kernel-based approaches.









\clearpage
\subsubsection*{Acknowledgements}
We thank Eric Strobl for kindly providing R-code for KCIT, RCIT and RCoT and Dino Sejdinovic for many helpful comments.

\bibliography{/home/jakobrunge/work/library}
\bibliographystyle{apalike}

\end{document}